\documentclass[sigconf,authorversion,nonacm]{acmart}

\AtBeginDocument{%
  }

\usepackage{subcaption}	
\usepackage{xcolor}

\newcommand{\qq}{\mathbf{q}}

\begin{document}

\title{A Theory of Stabilization by Skull Carving}

\author{Mathieu Lamarre}
\email{mlamarre@ea.com}
\affiliation{%
  \institution{SEED Electronic Arts}
  \city{Montréal}
  \state{Québec}
  \country{Canada}
}

\author{Patrick Anderson}
\email{panderson@ea.com}
\affiliation{%
  \institution{SEED Electronic Arts}
  \city{Montréal}
  \state{Québec}
  \country{Canada}
}

\author{Étienne Danvoye}
\email{edanvoye@ea.com}
\affiliation{%
  \institution{SEED Electronic Arts}
  \city{Montréal}
  \state{Québec}
  \country{Canada}
}

\begin{abstract}
Accurate stabilization of facial motion is essential for applications in photoreal avatar construction for 3D games, virtual reality, movies, and training data collection.
For the latter, stabilization must work automatically for the general population with people of varying morphology.
Distinguishing rigid skull motion from facial expressions is critical since misalignment between skull motion and facial expressions can lead to animation models that are hard to control and can not fit natural motion.
Existing methods struggle to work with sparse sets of very different expressions, such as when combining multiple units from the Facial Action Coding System (FACS).
Certain approaches are not robust enough, some depend on motion data to find stable points, while others make one-for-all invalid physiological assumptions.
In this paper, we leverage recent advances in neural signed distance fields and differentiable isosurface meshing to compute skull stabilization rigid transforms directly on unstructured triangle meshes or point clouds, significantly enhancing accuracy and robustness. 
We introduce the concept of a stable hull as the surface of the boolean intersection of stabilized scans, analogous to the visual hull in shape-from-silhouette and the photo hull from space carving. 
This hull resembles a skull overlaid with minimal soft tissue thickness, upper teeth are automatically included. 
Our skull carving algorithm simultaneously optimizes the stable hull shape and rigid transforms to get accurate stabilization of complex expressions for large diverse sets of people, outperforming existing methods.

\end{abstract}

\begin{teaserfigure}
\centering
\includegraphics[width=0.12\textwidth,keepaspectratio]{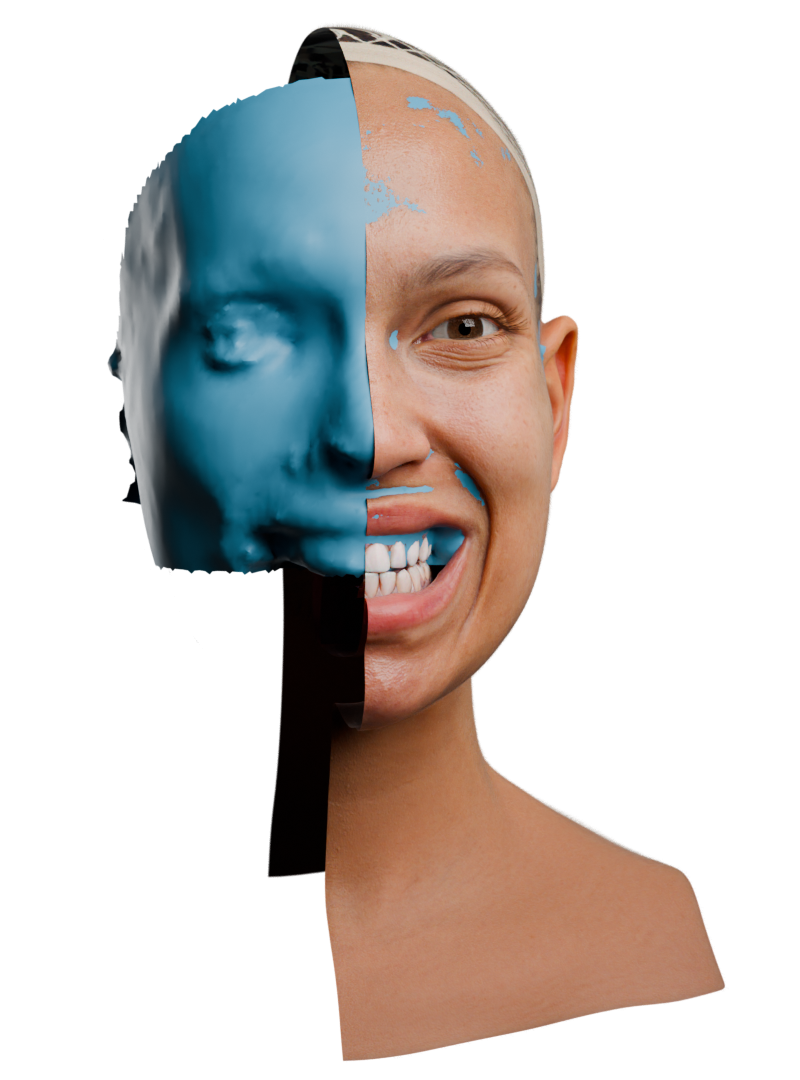}
\includegraphics[width=0.12\textwidth,keepaspectratio]{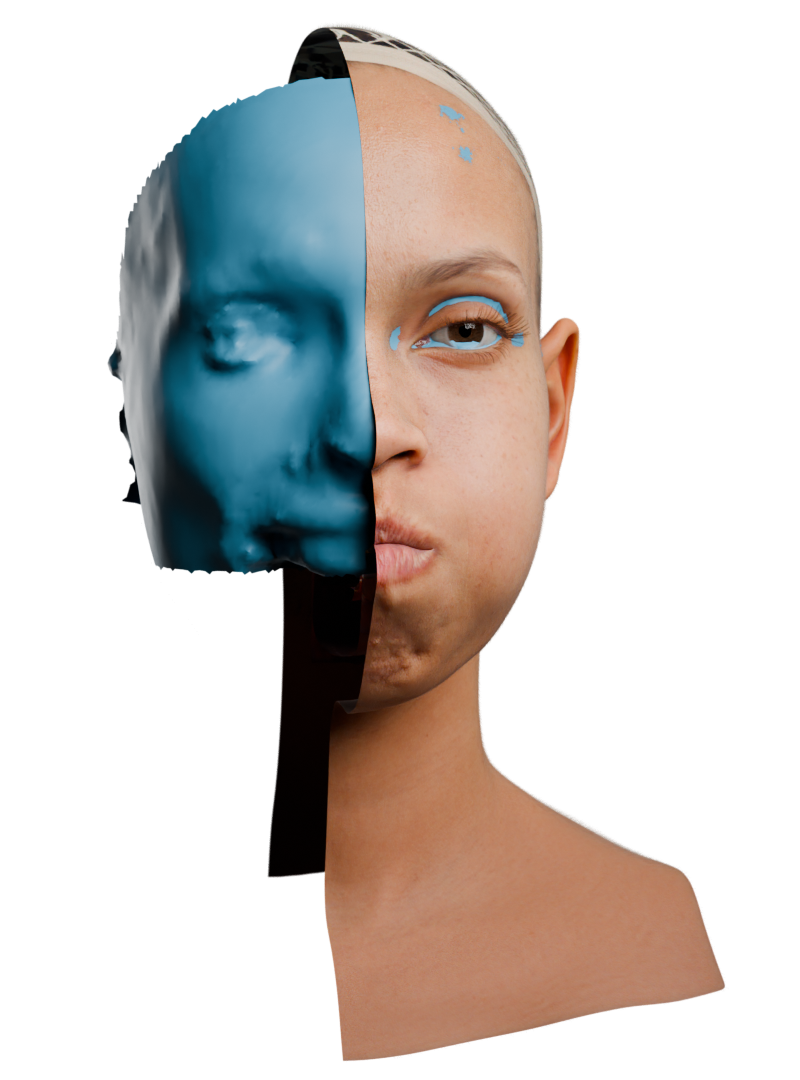}
\includegraphics[width=0.12\textwidth,keepaspectratio]{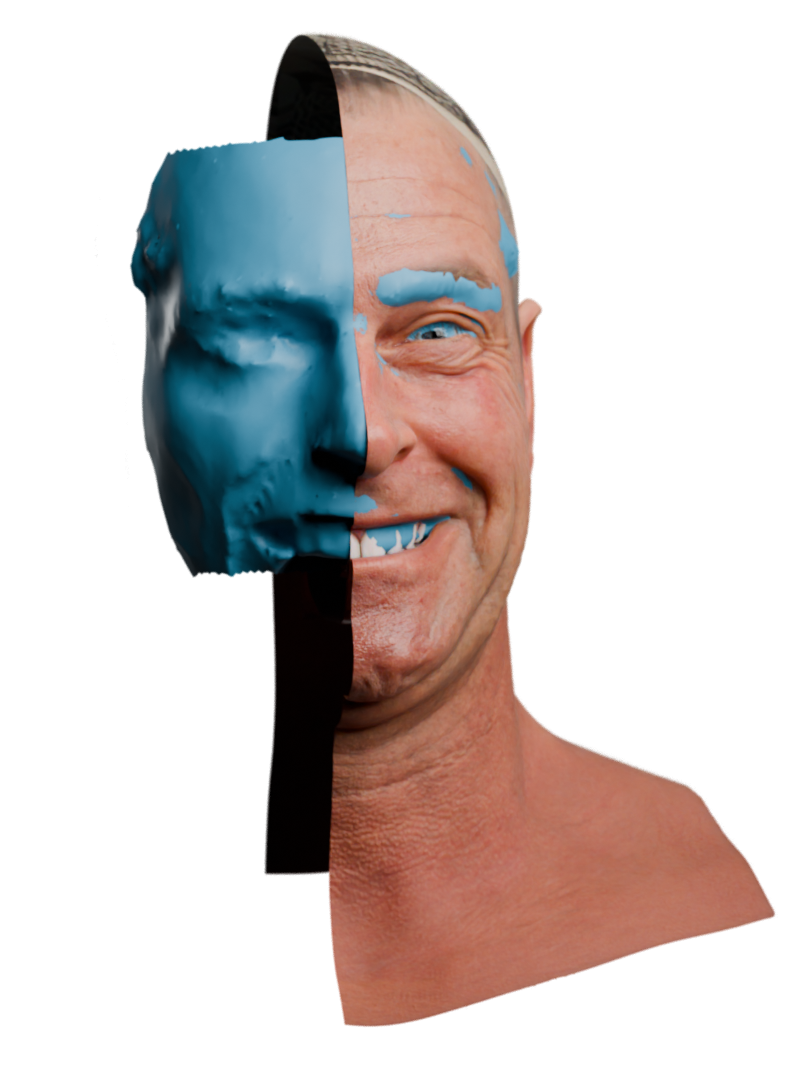}
\includegraphics[width=0.12\textwidth,keepaspectratio]{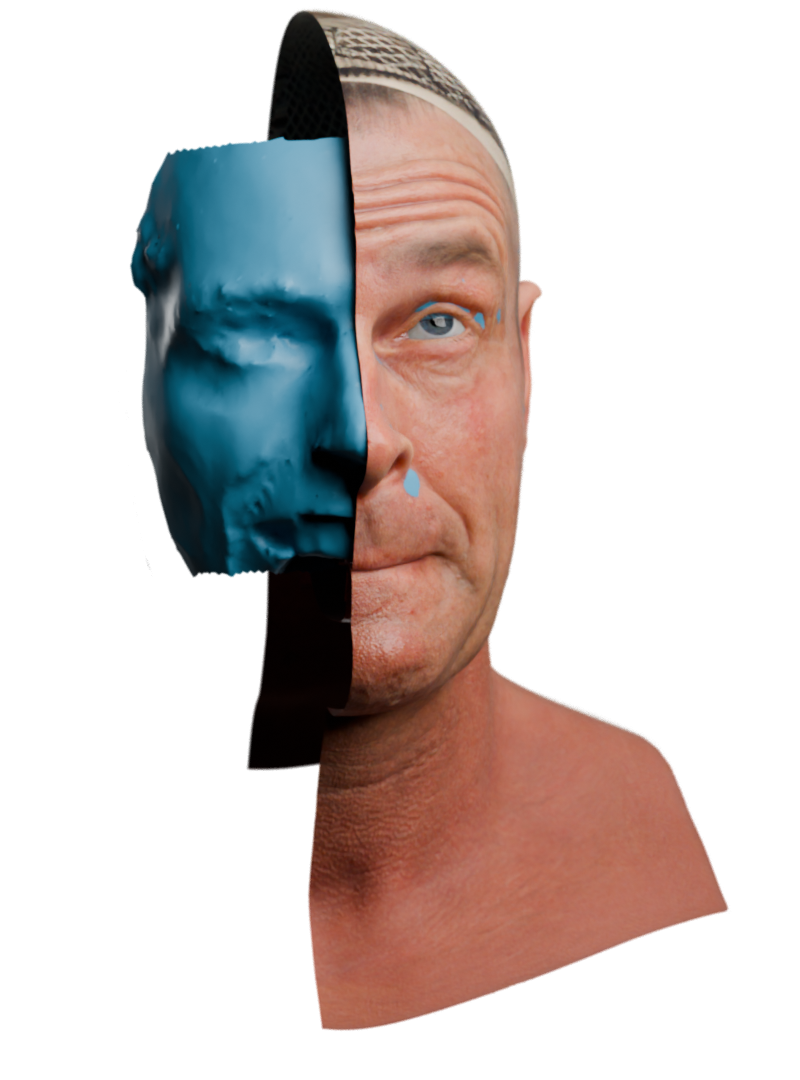}
\includegraphics[width=0.12\textwidth,keepaspectratio]{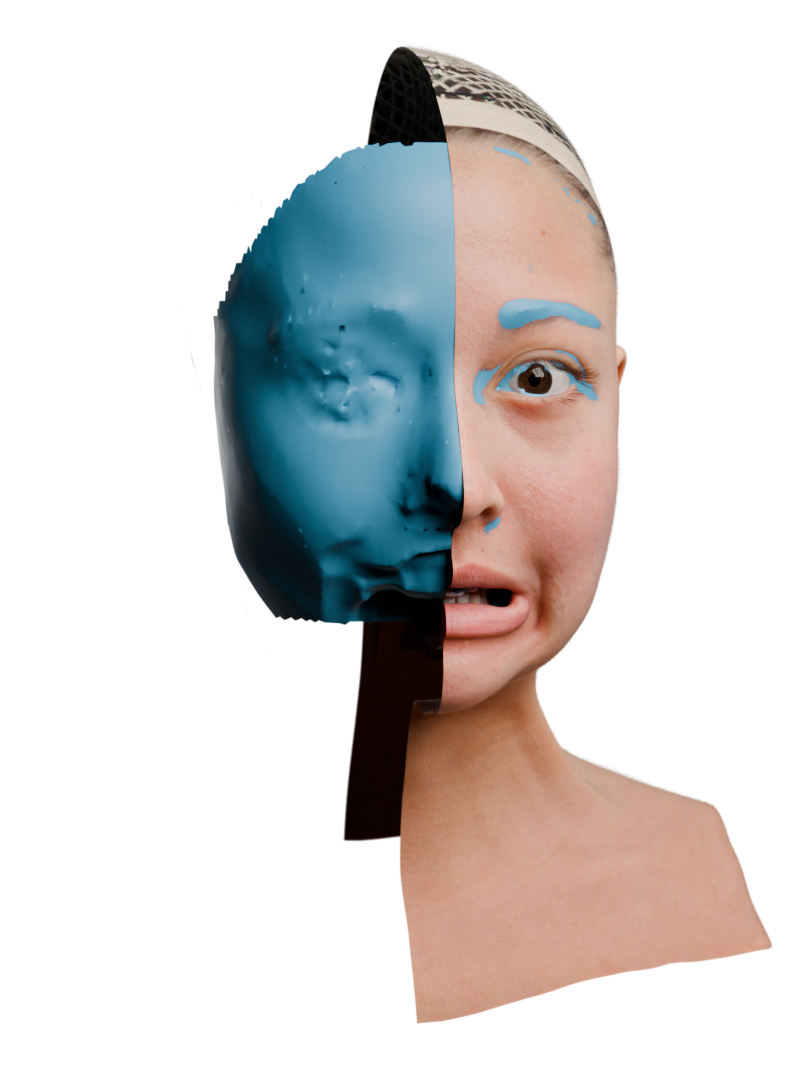}
\includegraphics[width=0.12\textwidth,keepaspectratio]{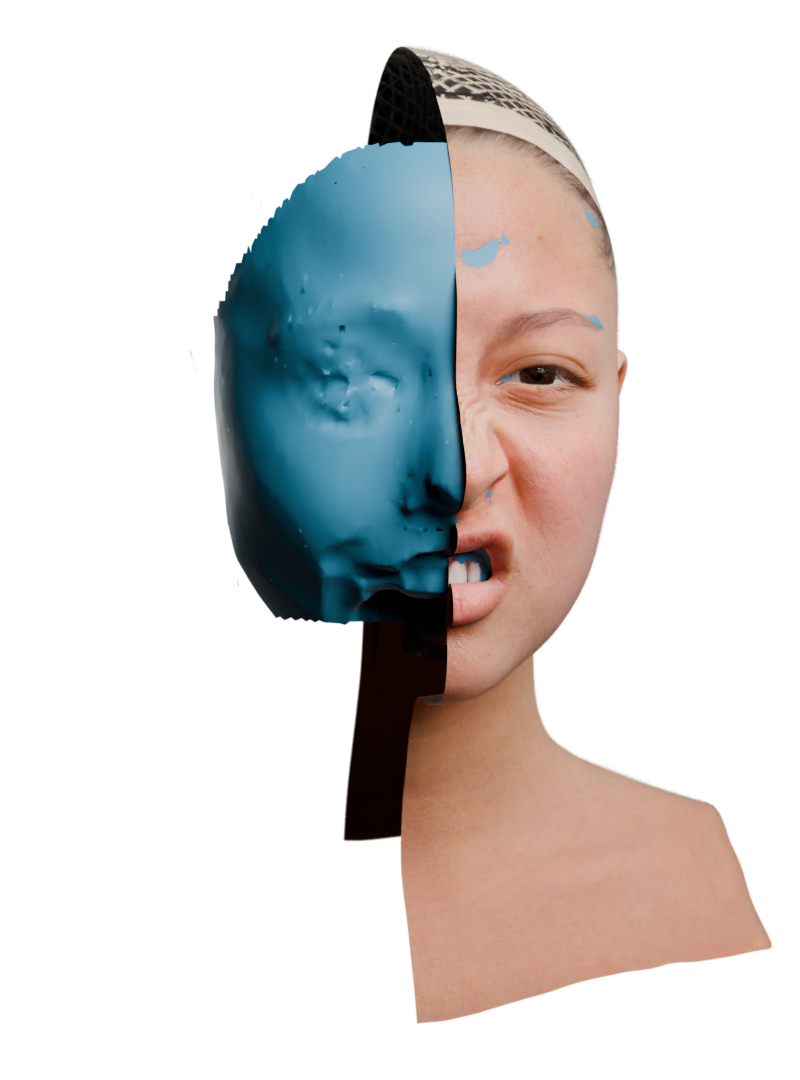}
\includegraphics[width=0.12\textwidth,keepaspectratio]{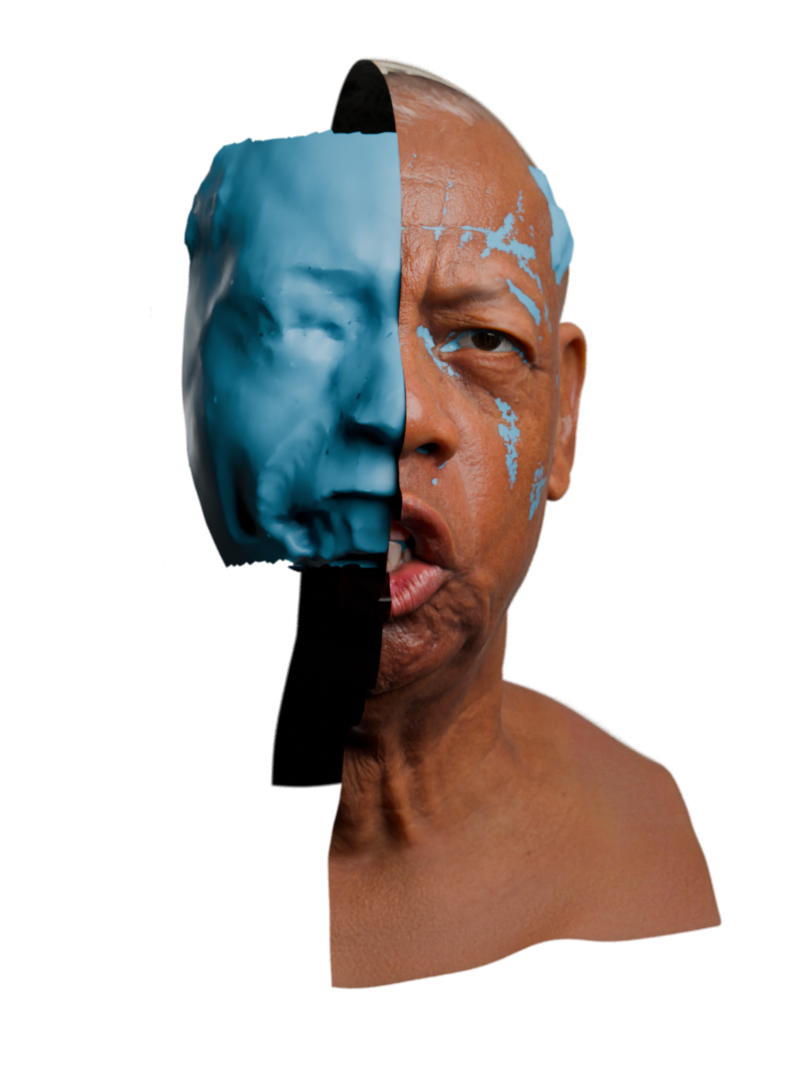}
\includegraphics[width=0.12\textwidth,keepaspectratio]{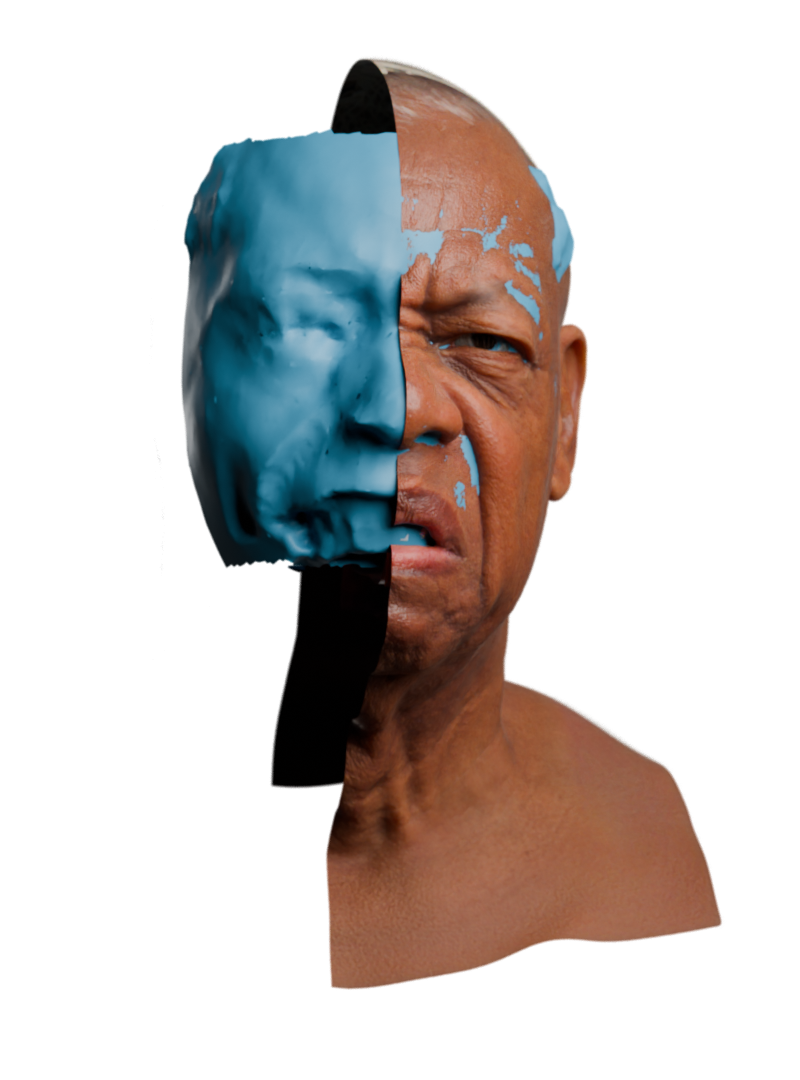}
\caption{We present the skull carving algorithm for skull stabilization in facial animation. Above a split-view of some stabilized expressions with the stable hull in blue.}
\label{fig:main_teaser_figure}
\Description[Paper teaser figure showing the concept of stable hull]{An image mosaic of five sub-images stacked horizontally, each with a different expression with the stable hull shown in a split mesh view}
\end{teaserfigure}

\maketitle

\section{Introduction}

\begin{figure}
\centering
    \includegraphics[width=0.22\linewidth,keepaspectratio]{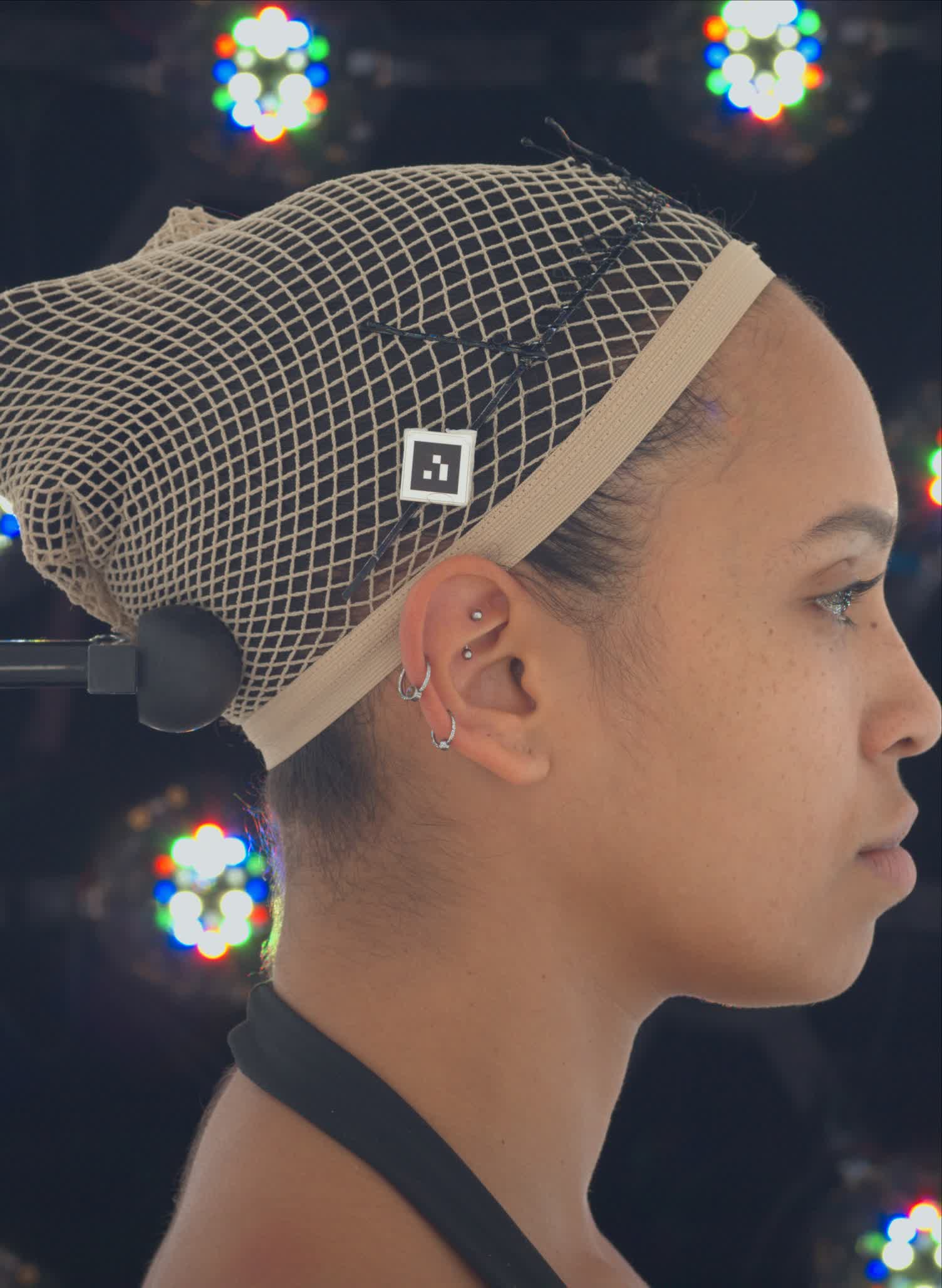}    
    \includegraphics[width=0.22\linewidth,keepaspectratio]{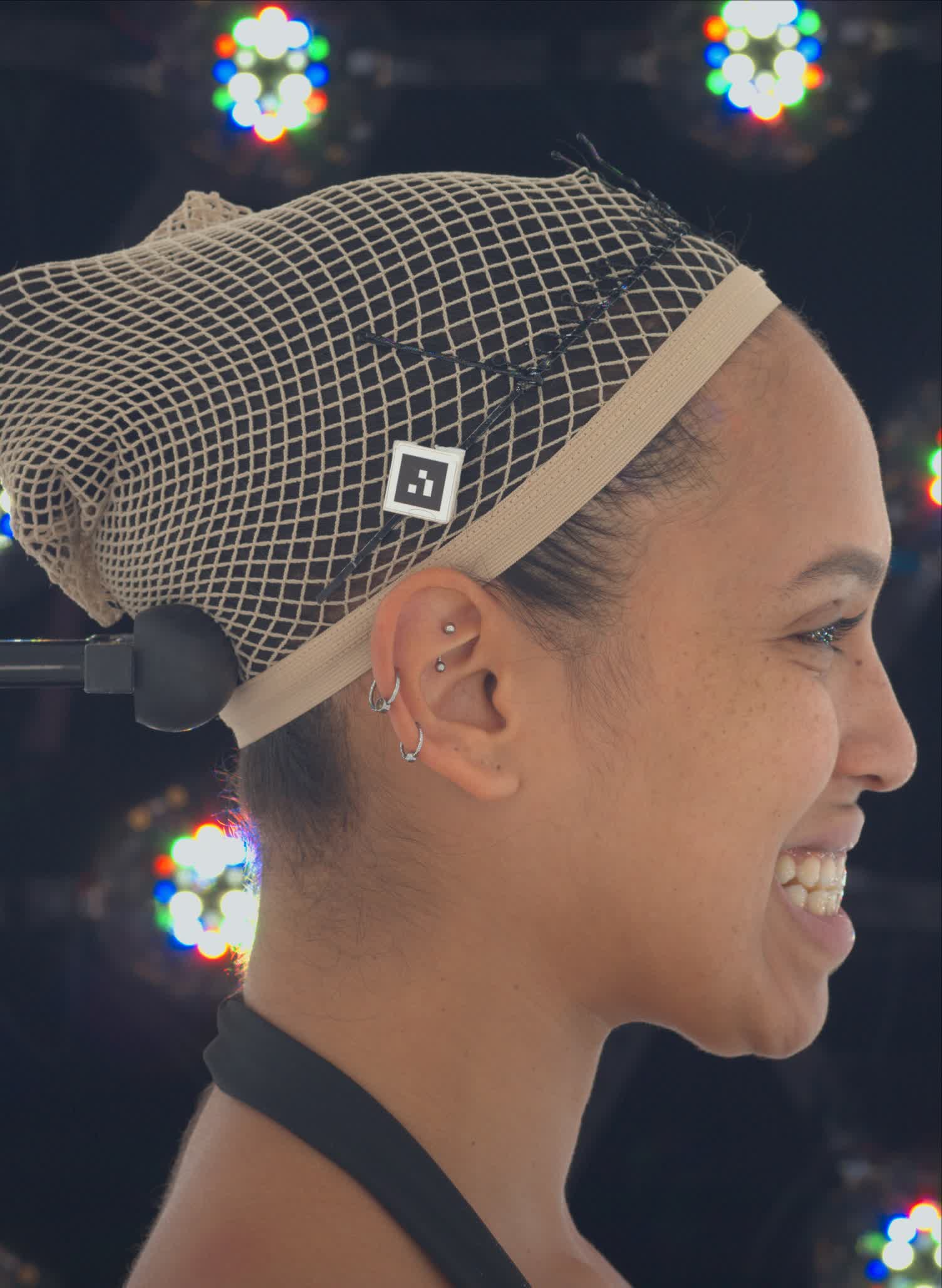}
    \includegraphics[width=0.22\linewidth,keepaspectratio]{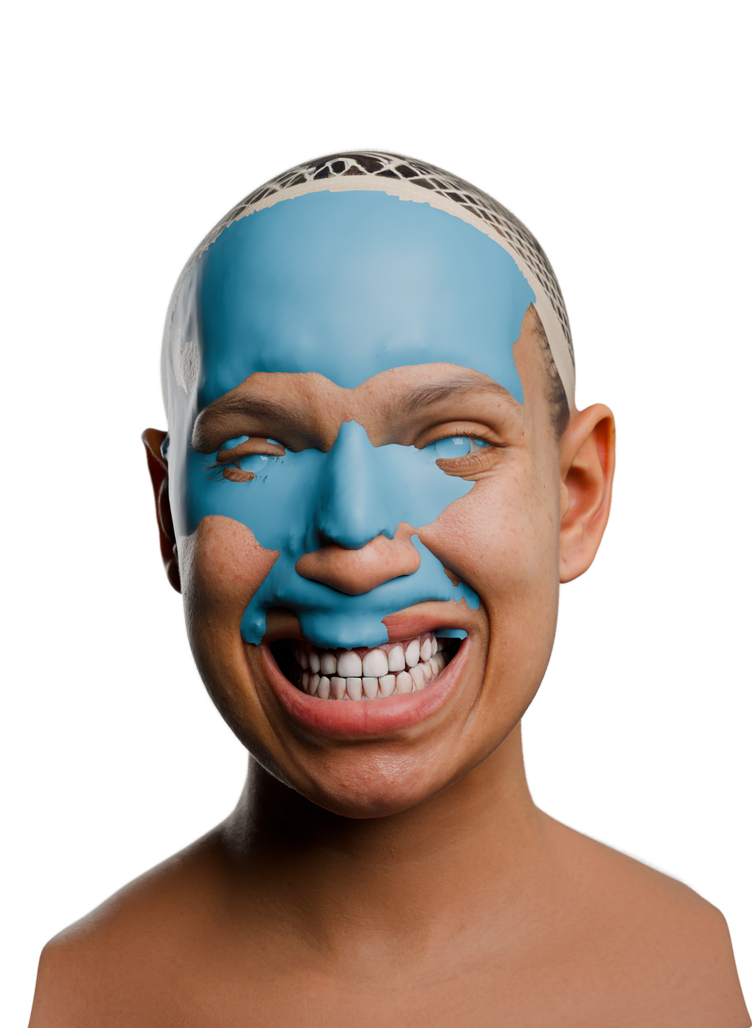}    
    \includegraphics[width=0.22\linewidth,keepaspectratio]{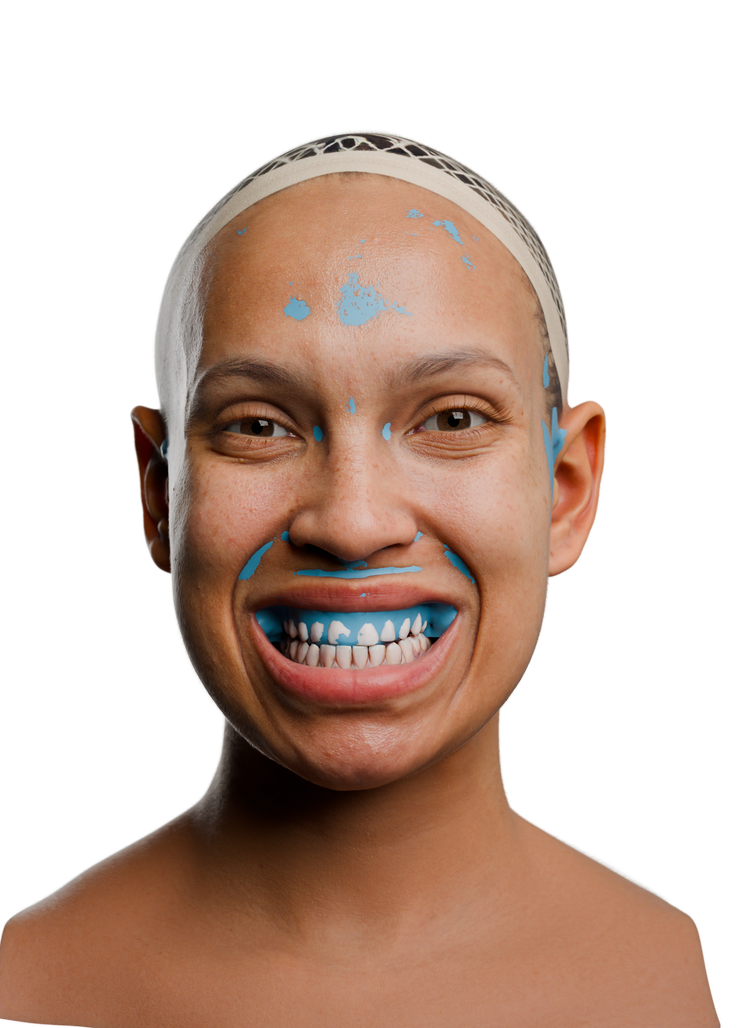}
\caption{Even with a headrest, the whole head moves when performing expressions (left). Stabilization is the process of estimating the rigid transform to remove the skull motion from non-rigid expression deformation (right).}
\label{fig:stabexplain}
\Description[Figure explaining stabilization]{The figure is made of 4 smaller pictures. The two on the left show a real-world picture of the capture rig with a girl neutral and AU10AU16AU25 expression with visible head motion. The two on the right show how stabilization starts with a large distance between the mesh and the skull and ends with the two well-aligned.}
\end{figure}

High-resolution photorealistic avatar likeness capture for hero characters in 3D games or virtual reality face-to-face conversation or digital double for movies is a mature but complex process that still requires significant manual effort.
On the other hand, major progress in the field of generative AI promises to make avatar
creation as simple as generating images from text. 
However, there is still some way to go before generated photoreal 3D facial animations become convincing enough for use in the previously mentioned applications.
See \cite{zollhofer2018state} for a review of the field,
\cite{saito2024rgca} for a recent example of a machine-learned photo-real animated avatar and \cite{wu2024text} for an example of text-to-avatar generation.
In any case, one of the best ways to improve both photoreal 3D avatar likeness capture and generative AI training data is to automate as many steps as possible in the creation pipeline while maintaining or improving the quality of the final result, especially for animations.

Current avatar animation models whether they are based on blendshapes or linear blend skinning
or more advanced machine learning methods first need the captured data to be normalized. 
The goal of likeness capture is to obtain a controllable model that is independent of the capturing modalities, for example, the head position and angle with respect to the rest of the body. 
A small but crucial step is to estimate the position and orientation of the skull. 
The skull must stay fixed when other control variables change, for example, opening the lips should not cause the upper teeth and eyes to move.
Strong facial expressions are often correlated with some head motion. 
Removing as much correlation between control variables
from the training data is desirable for next-generation neural network-based facial animation models.

If the goal of likeness capture is creating a database of heads representative of the general population and resources are limited, using static expressions with combined FACS units for a larger group of people instead of using 4D data from a smaller group is a good compromise.
This leads to the challenge of estimating the skull pose of various complex expressions from static captures. 
Existing methods require either 4D temporal data to find stable features or make simplifying assumptions about the skull shape and skin thickness, hence they are not appropriate for stabilizing an expression database including a variety of ethnicity, age, and body mass index (BMI), with a large difference in morphology and skin thickness. 

\section{Related Work}
\label{sec:related}

\cite{bouaziz2013online} uses Iterated Closest Point (ICP) with a custom template focusing on rigid regions of the upper face, like the forehead and the nose.
This method works relatively well for all expressions that do not involve strong upper face, e.g. eyebrows, motion.
However, it has limited accuracy because it relies on a single fixed template and the rest positions of points on the surface of the face have some variance.

\cite{derekbeelerStabilization} propose a non-linear optimization method that requires customizing a template skull model on each person by making several assumptions, for instance, that the skin thickness at five specific points on the face is constant between individuals.
This assumption doesn't hold at all.
For example, for the nose radix point between the eyes, Rhinoplasty literature \cite{dey2021predicting} shows that the standard deviation of the skin thickness at this point is 1.7 mm for a 6.7 mm average and that this value is correlated with the BMI.
Other assumptions include a fixed single one-for-all texture mask that describes how the skin thickness varies on the face
and a very simplified physics model for how the nose should bend.

The FLAME parametric facial animation model \cite{FLAME:SiggraphAsia2017} is trained on thousands of low-resolution laser scans
and thousands of frames of 4D data.
It includes a linear blend skinning model with a bone for the skull that may be used for stabilization.
However, the model is trained on raw scan data and the minimized energy function has no skull-specific term, so the fitting error is distributed randomly between shape, expression, and pose parameters and the skull pose will carry some inaccuracy.

\cite{lamarre2018face} observe that in the skull coordinate frame, the zero distance-to-neutral mode of the head mesh vertex histogram is maximized.
Their mode pursuit method searches for the skull pose trajectory that minimizes a custom loss function that gradually approximates the $\ell_0$ norm of vertex positions and velocities.
This is equivalent to maximizing the zero mode of the distance-to-neutral and velocity histograms.
Since mode-pursuit is carried on both positions and velocities of the head mesh vertices, without temporal data, on static expressions, this method lacks motion information to find stable regions.
It reduces to an approximate $\ell_0$ norm ICP, similar to \cite{bouaziz2013sparseicp} but with a better-behaved penalty function for gradient descent.
Mode pursuit fails if the maximum zero mode hypothesis fails, which occurs when all points on the upper portion of the head move at the same time.
This happens when many FACS action units are combined.

\section{Method}
\label{sec:method}

\begin{figure*}
\centering
\includegraphics[width=\textwidth,keepaspectratio]{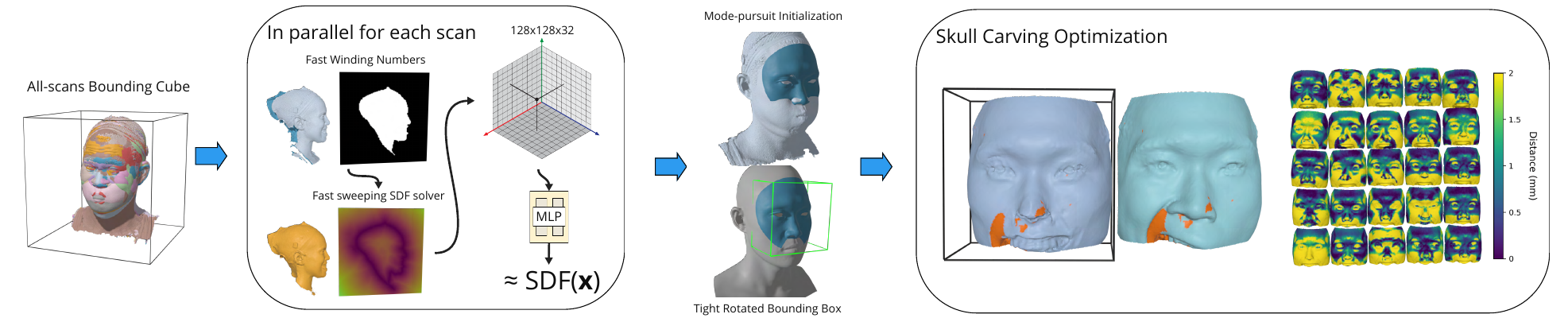}
\caption{Overview of the method: obtain compact SDF representations for each scan, initialize stabilization rigid transforms with mode-pursuit, finally optimize the stable hull and transforms simultaneously minimizing the zero mode to each SDF.}
\label{fig:overview}
\Description[Overview of the skull carving method]{Overview of the method: obtain compact SDF representations for each scan, initialize stabilization transforms with mode-pursuit, finally optimize the stable hull and rigid transforms simultaneously minimizing the zero distance mode to each SDF.}
\end{figure*}

Our skull carving method works on a set of 3D facial static expression scans.
It doesn't depend on a specific capture method and only requires a 3D point cloud or unstructured triangle mesh per expression and a template mesh aligned to the neutral to be used as the reference coordinate frame;
see \cite{zollhofer2018state} for a survey of methods to create 3D facial scans.

To explain our algorithm we first need to explain the stable hull concept, which is analogous to two well-known ideas in computer vision: the visual and photo hulls.
The visual hull introduced by \cite{laurentini1994visual} is the intersection of the binary silhouettes perspective cones of a set of cameras. 
\cite{kutulakos2000theory} invented the skull carving algorithm and the photo hull concept, which is the remaining isosurface after carving voxels with a photo-consistency test.
We named our algorithm after theirs. 

Given that each expression scan is converted to a signed distance field (SDF), with a defined interior (negative) and exterior (positive), when each SDF is rigidly transformed to the stabilized skull coordinate frame, the zero isosurface of the intersection of their interior volume is the stable hull.
The optimal stable hull of a sufficiently large set of expressions should be the skull layered with the minimum observable soft tissue thickness for the person.
If an expression shows visible upper teeth, the stable hull will fit them accurately.

Our main hypothesis is that \textit{for each expression, there is always a large enough region of the stable hull that is close enough to the scan surface to maximize the zero distance mode in the skull coordinate frame.}

It is true for all expressions where upper teeth are sufficiently visible on the scan surface since they are part of the stable hull and also on the scan surface.
For the other expressions, we assume that muscle and skin move around the face and can never totally cover the stable hull.
This hypothesis is enough to build an energy function that can be minimized to solve simultaneously for the skull poses and stable hull using gradient descent.
Using the mode as the main loss function makes the method robust, support regions may be sparse and small.

\subsection{Raw 3D Scan to SDF}

To convert raw 3D scans to SDFs, we first compute the bounding cube of all scans and create voxel grids for each scan with the same world coordinate extent.
Next, we determine if each voxel is inside or outside the scan mesh using the Fast Winding Number method of \cite{fast_winding_numbers}.
In practice, we only tested with unstructured triangle mesh input, but this algorithm also works on 3D point clouds.
SDFs are then computed with the Eikonal equation solver of \cite{fast_sweep} which only works on cubic voxel grids.
To implement an efficient stabilization process, scan SDF must be compact and fast to evaluate on the graphical processing unit (GPU).
The expression sets being stabilized may have between 10 and 100 scans.
Voxel grid evaluations are very fast but even at low bit depth they require too much memory to stabilize 100 expressions.
We propose to use a very simple tri-plane-based neural SDF model similar, but simpler, to \cite{triplane}.
For our purpose, feeding the output of the 128x128x32 tri-plane features to a single multi-layer perceptron (MLP) with 2 hidden layers of 196 neurons and ReLu activation is enough to get sub-millimeter accuracy close to the scan surface.
We train the distinct parameters $\theta_i$ of this model to approximate each expression $SDF_i$ over the whole voxel domain $\Omega$.
\begin{equation}
    \phi_{\theta_i}(x) \approx SDF_i(x), \forall x \in \Omega
\end{equation}
This model evaluates fast and takes 7 MB of GPU memory per scan.

\subsection{Skull Carving Optimization}

Skull carving is a non-linear optimization problem solved with gradient descent.
To model rigid transformations, we use unit dual quaternions which are well suited for numerical optimization.
Carving is implemented by taking the maximum distance over all stabilized expressions.
If for a point in stabilized space, the maximum distance to any expression is positive, this point is considered outside the stable volume.
Let $\gamma$ be the differentiable isosurface extraction function of Flexicube \cite{flexicube}.
The function $\gamma$ takes a scalar field as input and outputs the vertex positions of the stable hull triangle mesh.
In practice, Flexicube also outputs the triangle vertex indices which are useful for visualization but are not used during optimization.
Let $Q = \{\qq_1 = I, \qq_i \}_{i=2}^{N}$ be the set of stabilization dual quaternions and $X_r$ be an array of voxel grid points in the neutral reference frame.
The first transform is identity and is matched with the neutral reference SDF.

The stable~hull function in the reference frame $\mathcal{S}$ is
\begin{equation}
    \mathcal{S}(Q) = \gamma \Big( \max_{i \in [N]} \phi_{\theta_i} (\qq_{i} X_r \overline{\qq_{i}}) \Big)
\label{stablehulleq}
\end{equation}
With $\psi$ as the $\ell_0$ norm approximating penalty function of \cite{lamarre2018face}, the optimization process is
\begin{equation}
    \arg\min_{Q} \frac{1}{N} \sum_{i=1}^{N} \psi \Bigg( \phi_{\theta_i} \Big( \qq_{i} \mathcal{S}(Q) \overline{\qq_{i}}  \Big) \Bigg)
\label{optimeq}
\end{equation}
These equations are implemented in Pytorch and optimized using the Adam optimizer.
We implement a two-step mode-pursuit schedule, first optimizing with a histogram bin size of 2 mm and then 1 mm.
The voxel grid point set $X_r$ size is $40^3$ but we use a mask to ignore voxels very far from inside or outside all surfaces at the initialization stage (+/- 4 mm) to accelerate computations.
The masked voxels have fixed signed distances and do not influence the stable hull mesh.
Each step runs for 2000 iterations with a learning rate set to $1\mathrm{e}{-3}$.
For 25 expressions, the process takes 5 minutes on a GeForce GTX 3090 and requires 15 GB of GPU memory.

\subsection{Initialization}

Initialization is important as we found the energy function to be non-convex.
Since the head can move by more than 2 cm, a coarse head alignment method using face features from Mediapipe \cite{mediapipe} is used first.
Afterward, we apply the mode pursuit method of \cite{lamarre2018face} using SDF distances instead of vertex-vertex matches.

\section{Results}
\label{sec:results}

To evaluate the skull carving algorithm on a significant sample of the population, we build a database of 32 persons.
We picked randomly from our larger internal head capture database to get an equal partition of male-female, asian-black-latino-white phenotype, and young-old. 
The sample contains people with different BMIs.

For comparison with our skull carving algorithm, we test five stabilization methods: (FLAME, $\ell_2$-LBFGS-ICP, $\ell_1$-LBFGS-ICP, GM-LBFGS-ICP, mode pursuit).
For FLAME we use the official Chumpy-based fitting implementation \cite{FLAME:SiggraphAsia2017}. 
The stabilization rigid transform is computed with the pose of the second bone in the FLAME skinning hierarchy.
We achieved the best results by initially fitting the FLAME model to the neutral scan with expression parameter estimation disabled to obtain the shape parameters.
Subsequently, we fitted each expression scan using these fixed shape parameters while enabling the expression parameters.
Since we have neural SDFs available for which the gradient of the distance is available, we can use a gradient descent version of ICP which was shown in Levenberg-Marquardt ICP (LM-ICP) \cite{fitzgibbon2003robust} to be more flexible and robust than the standard implementation.
LM is not available in Pytorch but L-BFGS with strong Wolfe line search is, so we implemented L-BFGS ICP with three different loss functions: $\ell_2$, $\ell_1$ and Geman-McClure (GM) robust function as in \cite{FLAME:SiggraphAsia2017}.
ICP and mode-pursuit use the mask illustrated in Fig.\ref{fig:overview}, which is also used as the basis to compute the skull carving bounding box.
As an ablation study, we also test a version of skull carving where surface extraction gradients and grid deformation offsets are disabled. 

The quantitative evaluation process is manual and performed on expressions with visible upper teeth. 
There is a total of 227 such expression scans, between 4 and 10 per person with an average of 7. 
The expression list varies slightly between captures, but even for the same action unit combo the visibility of the upper teeth may vary because of morphology. 
The results are compiled in Table~\ref{table:teeth_alignement_error}. 
Skull carving is superior to all other methods.
All ICP methods have similar error counts below 2 mm. 
Robust estimators do not work well on this dataset because they contain combined action units that deform all the upper part of the face, meaning the outlier percentage is too high for these estimators to work.
Mode pursuit often fails badly for the same reason.
FLAME results are generally consistent but with a relatively high error for this task.
It seems to indicate that training a similar model with a specific algorithm for skull pose estimation could be beneficial.
Skull carving without a differentiable surface extraction method is worse than FLAME and similar to ICP methods.
Surface extraction gradients make skull carving an expression global method.
When the gradients are disabled, skull carving will converge to a local minima focusing on an expression subset which depends on the initialization and the shape of the initial stable hull.

\begin{table}[ht]
\centering
\caption{In a 3D DCC software, for each of the 32 persons in the sample, for the set of expressions with visible upper teeth, we align an upper teeth template manually on the stabilized expression scans and estimate the worst-case translation error in the set and sort each person between four error brackets.}
\begin{tabular}{ccccc}
\hline
Method             & \multicolumn{4}{c}{Max Upper Teeth Alignment Error} \\ 
                   & <= 1 mm & <= 2 mm & <= 3 mm & > 3 mm \\ 
                   & (\%) & (\%) & (\%) & (\%) \\ \hline
$\ell_2$-ICP       & 3  & 22 & 88 & 12 \\
$\ell_1$-ICP       & 3  & 28 & 63 & 37\\
GM-ICP             & 6  & 25 & 56 & 44 \\
FLAME              & 9  & 50 & 94  & 6 \\
Mode pursuit       & 13 & 38 & 53  & 47 \\
\textbf{Skull Carving}  & \textbf{78} & \textbf{97} & \textbf{97}  & 3      \\ \hline
\textit{NG\footnotemark Skull Carving}   & 13  & 31 & 63 & 38  \\ \hline
\end{tabular}
\label{table:teeth_alignement_error}
\end{table}

\footnotetext{No surface extraction gradient}

\section{Discussion and Conclusion}
\label{sec:discussion}

Our experiment shows that differentiable isosurface extraction is a necessary component of the skull carving stabilization algorithm.
We observed that Eq. \eqref{optimeq} does not converge from a coarse initialization.
It seems the stable hull must have already well-defined 3D features like teeth stub and a fairly complete nose arch for convergence.
We hypothesize that adding a domain-specific regularization term on the stable hull shape, for example, a distance to a learned subspace, or a statistical anthropometric distance, would improve convergence.
Skull carving is not adapted to 4D data stabilization since even on a high-end 48 GB GPU, it can process only about 100 frames at a time. 
More research is needed to use the stable hull concept on a large number of sequential scans.

In this paper, we introduced a new solution to the facial animation skull stabilization problem on a sparse set of combined FACS unit expressions.
This problem is more challenging than stabilization on 4D capture since there is no motion information available.
While there is room for improvement, the skull carving algorithm is more accurate than existing methods on our diverse database and we believe it should generalize well because it makes no morphological assumption.

\bibliographystyle{ACM-Reference-Format}
\bibliography{skull_carving}


\begin{thebibliography}{17}


\ifx \showCODEN    \undefined \def \showCODEN     #1{\unskip}     \fi
\ifx \showDOI      \undefined \def \showDOI       #1{#1}\fi
\ifx \showISBNx    \undefined \def \showISBNx     #1{\unskip}     \fi
\ifx \showISBNxiii \undefined \def \showISBNxiii  #1{\unskip}     \fi
\ifx \showISSN     \undefined \def \showISSN      #1{\unskip}     \fi
\ifx \showLCCN     \undefined \def \showLCCN      #1{\unskip}     \fi
\ifx \shownote     \undefined \def \shownote      #1{#1}          \fi
\ifx \showarticletitle \undefined \def \showarticletitle #1{#1}   \fi
\ifx \showURL      \undefined \def \showURL       {\relax}        \fi
\providecommand\bibfield[2]{#2}
\providecommand\bibinfo[2]{#2}
\providecommand\natexlab[1]{#1}
\providecommand\showeprint[2][]{arXiv:#2}

\bibitem[Barill et~al\mbox{.}(2018)]%
        {fast_winding_numbers}
\bibfield{author}{\bibinfo{person}{Gavin Barill}, \bibinfo{person}{Nia Dickson}, \bibinfo{person}{Ryan Schmidt}, \bibinfo{person}{David~I.W. Levin}, {and} \bibinfo{person}{Alec Jacobson}.} \bibinfo{year}{2018}\natexlab{}.
\newblock \showarticletitle{Fast Winding Numbers for Soups and Clouds}.
\newblock \bibinfo{journal}{\emph{ACM Transactions on Graphics}} (\bibinfo{year}{2018}).
\newblock


\bibitem[Beeler and Bradley(2014)]%
        {derekbeelerStabilization}
\bibfield{author}{\bibinfo{person}{Thabo Beeler} {and} \bibinfo{person}{Derek Bradley}.} \bibinfo{year}{2014}\natexlab{}.
\newblock \showarticletitle{Rigid Stabilization of Facial Expressions}.
\newblock \bibinfo{journal}{\emph{ACM Trans. Graph.}} \bibinfo{volume}{33}, \bibinfo{number}{4}, Article \bibinfo{articleno}{44} (\bibinfo{date}{July} \bibinfo{year}{2014}), \bibinfo{numpages}{9}~pages.
\newblock
\showISSN{0730-0301}


\bibitem[Bouaziz et~al\mbox{.}(2013a)]%
        {bouaziz2013sparseicp}
\bibfield{author}{\bibinfo{person}{Sofien Bouaziz}, \bibinfo{person}{Andrea Tagliasacchi}, {and} \bibinfo{person}{Mark Pauly}.} \bibinfo{year}{2013}\natexlab{a}.
\newblock \showarticletitle{Sparse Iterative Closest Point}. In \bibinfo{booktitle}{\emph{Eurographics/ACMSIGGRAPH SGP}} \emph{(\bibinfo{series}{SGP '13})}. \bibinfo{pages}{113--123}.
\newblock


\bibitem[Bouaziz et~al\mbox{.}(2013b)]%
        {bouaziz2013online}
\bibfield{author}{\bibinfo{person}{Sofien Bouaziz}, \bibinfo{person}{Yangang Wang}, {and} \bibinfo{person}{Mark Pauly}.} \bibinfo{year}{2013}\natexlab{b}.
\newblock \showarticletitle{Online modeling for realtime facial animation}.
\newblock \bibinfo{journal}{\emph{ACM Transactions on Graphics (ToG)}} \bibinfo{volume}{32}, \bibinfo{number}{4} (\bibinfo{year}{2013}), \bibinfo{pages}{1--10}.
\newblock


\bibitem[Dey et~al\mbox{.}(2021)]%
        {dey2021predicting}
\bibfield{author}{\bibinfo{person}{Jacob~K Dey}, \bibinfo{person}{Chelsey~A Recker}, \bibinfo{person}{Michael~D Olson}, \bibinfo{person}{Andrew~J Bowen}, {and} \bibinfo{person}{Grant~S Hamilton~III}.} \bibinfo{year}{2021}\natexlab{}.
\newblock \showarticletitle{Predicting nasal soft tissue envelope thickness for rhinoplasty: a model based on visual examination of the nose}.
\newblock \bibinfo{journal}{\emph{Annals of Otology, Rhinology \& Laryngology}} \bibinfo{volume}{130}, \bibinfo{number}{1} (\bibinfo{year}{2021}), \bibinfo{pages}{60--66}.
\newblock


\bibitem[Fitzgibbon(2003)]%
        {fitzgibbon2003robust}
\bibfield{author}{\bibinfo{person}{Andrew~W Fitzgibbon}.} \bibinfo{year}{2003}\natexlab{}.
\newblock \showarticletitle{Robust registration of 2D and 3D point sets}.
\newblock \bibinfo{journal}{\emph{Image and vision computing}} \bibinfo{volume}{21}, \bibinfo{number}{13-14} (\bibinfo{year}{2003}), \bibinfo{pages}{1145--1153}.
\newblock


\bibitem[Kutulakos and Seitz(2000)]%
        {kutulakos2000theory}
\bibfield{author}{\bibinfo{person}{Kiriakos~N Kutulakos} {and} \bibinfo{person}{Steven~M Seitz}.} \bibinfo{year}{2000}\natexlab{}.
\newblock \showarticletitle{A theory of shape by space carving}.
\newblock \bibinfo{journal}{\emph{International journal of computer vision}}  \bibinfo{volume}{38} (\bibinfo{year}{2000}), \bibinfo{pages}{199--218}.
\newblock


\bibitem[Lamarre et~al\mbox{.}(2018)]%
        {lamarre2018face}
\bibfield{author}{\bibinfo{person}{Mathieu Lamarre}, \bibinfo{person}{John~P Lewis}, {and} \bibinfo{person}{Etienne Danvoye}.} \bibinfo{year}{2018}\natexlab{}.
\newblock \showarticletitle{Face stabilization by mode pursuit for avatar construction}. In \bibinfo{booktitle}{\emph{2018 IVCNZ}}. IEEE, \bibinfo{pages}{1--6}.
\newblock


\bibitem[Laurentini(1994)]%
        {laurentini1994visual}
\bibfield{author}{\bibinfo{person}{Aldo Laurentini}.} \bibinfo{year}{1994}\natexlab{}.
\newblock \showarticletitle{The visual hull concept for silhouette-based image understanding}.
\newblock \bibinfo{journal}{\emph{IEEE PAMI}} \bibinfo{volume}{16}, \bibinfo{number}{2} (\bibinfo{year}{1994}), \bibinfo{pages}{150--162}.
\newblock


\bibitem[Li et~al\mbox{.}(2017)]%
        {FLAME:SiggraphAsia2017}
\bibfield{author}{\bibinfo{person}{Tianye Li}, \bibinfo{person}{Timo Bolkart}, \bibinfo{person}{Michael.~J. Black}, \bibinfo{person}{Hao Li}, {and} \bibinfo{person}{Javier Romero}.} \bibinfo{year}{2017}\natexlab{}.
\newblock \showarticletitle{Learning a model of facial shape and expression from {4D} scans}.
\newblock \bibinfo{journal}{\emph{ACM Transactions on Graphics, (Proc. SIGGRAPH Asia)}} \bibinfo{volume}{36}, \bibinfo{number}{6} (\bibinfo{year}{2017}), \bibinfo{pages}{194:1--194:17}.
\newblock


\bibitem[Lugaresi et~al\mbox{.}(2019)]%
        {mediapipe}
\bibfield{author}{\bibinfo{person}{Camillo Lugaresi}, \bibinfo{person}{Jiuqiang Tang}, {and} \bibinfo{person}{Hadon Nash}.} \bibinfo{year}{2019}\natexlab{}.
\newblock \showarticletitle{MediaPipe: A Framework for Perceiving and Processing Reality}. In \bibinfo{booktitle}{\emph{Third Workshop on Computer Vision for AR/VR CVPR 2019}}.
\newblock


\bibitem[Saito et~al\mbox{.}(2024)]%
        {saito2024rgca}
\bibfield{author}{\bibinfo{person}{Shunsuke Saito}, \bibinfo{person}{Gabriel Schwartz}, \bibinfo{person}{Tomas Simon}, \bibinfo{person}{Junxuan Li}, {and} \bibinfo{person}{Giljoo Nam}.} \bibinfo{year}{2024}\natexlab{}.
\newblock \showarticletitle{Relightable Gaussian Codec Avatars}. In \bibinfo{booktitle}{\emph{CVPR}}.
\newblock


\bibitem[Shen et~al\mbox{.}(2023)]%
        {flexicube}
\bibfield{author}{\bibinfo{person}{Tianchang Shen}, \bibinfo{person}{Jacob Munkberg}, \bibinfo{person}{Jon Hasselgren}, \bibinfo{person}{Kangxue Yin}, \bibinfo{person}{Zian Wang}, \bibinfo{person}{Wenzheng Chen}, \bibinfo{person}{Zan Gojcic}, \bibinfo{person}{Sanja Fidler}, \bibinfo{person}{Nicholas Sharp}, {and} \bibinfo{person}{Jun Gao}.} \bibinfo{year}{2023}\natexlab{}.
\newblock \showarticletitle{Flexible Isosurface Extraction for Gradient-Based Mesh Optimization}.
\newblock \bibinfo{journal}{\emph{ACM Trans. Graph.}} \bibinfo{volume}{42}, \bibinfo{number}{4}, Article \bibinfo{articleno}{37} (\bibinfo{date}{jul} \bibinfo{year}{2023}), \bibinfo{numpages}{16}~pages.
\newblock
\showISSN{0730-0301}
\urldef\tempurl%
\url{https://doi.org/10.1145/3592430}
\showDOI{\tempurl}


\bibitem[Vicini et~al\mbox{.}(2022)]%
        {fast_sweep}
\bibfield{author}{\bibinfo{person}{Delio Vicini}, \bibinfo{person}{Sébastien Speierer}, {and} \bibinfo{person}{Wenzel Jakob}.} \bibinfo{year}{2022}\natexlab{}.
\newblock \showarticletitle{Differentiable Signed Distance Function Rendering}.
\newblock \bibinfo{journal}{\emph{Transactions on Graphics (Proceedings of SIGGRAPH)}} \bibinfo{volume}{41}, \bibinfo{number}{4} (\bibinfo{date}{July} \bibinfo{year}{2022}), \bibinfo{pages}{125:1--125:18}.
\newblock
\urldef\tempurl%
\url{https://doi.org/10.1145/3528223.3530139}
\showDOI{\tempurl}


\bibitem[Wang et~al\mbox{.}(2023)]%
        {triplane}
\bibfield{author}{\bibinfo{person}{Yiqun Wang}, \bibinfo{person}{Ivan Skorokhodov}, {and} \bibinfo{person}{Peter Wonka}.} \bibinfo{year}{2023}\natexlab{}.
\newblock \showarticletitle{Pet-neus: Positional encoding tri-planes for neural surfaces}. In \bibinfo{booktitle}{\emph{Proceedings of the IEEE/CVF Conference on Computer Vision and Pattern Recognition}}. \bibinfo{pages}{12598--12607}.
\newblock


\bibitem[Wu et~al\mbox{.}(2024)]%
        {wu2024text}
\bibfield{author}{\bibinfo{person}{Yunjie Wu}, \bibinfo{person}{Yapeng Meng}, \bibinfo{person}{Zhipeng Hu}, \bibinfo{person}{Lincheng Li}, \bibinfo{person}{Haoqian Wu}, \bibinfo{person}{Kun Zhou}, \bibinfo{person}{Weiwei Xu}, {and} \bibinfo{person}{Xin Yu}.} \bibinfo{year}{2024}\natexlab{}.
\newblock \showarticletitle{Text-Guided 3D Face Synthesis-From Generation to Editing}. In \bibinfo{booktitle}{\emph{Proceedings of the IEEE/CVF CVPR}}. \bibinfo{pages}{1260--1269}.
\newblock


\bibitem[Zollh{\"o}fer et~al\mbox{.}(2018)]%
        {zollhofer2018state}
\bibfield{author}{\bibinfo{person}{Michael Zollh{\"o}fer}, \bibinfo{person}{Justus Thies}, \bibinfo{person}{Pablo Garrido}, \bibinfo{person}{Derek Bradley}, \bibinfo{person}{Thabo Beeler}, \bibinfo{person}{Patrick P{\'e}rez}, \bibinfo{person}{Marc Stamminger}, \bibinfo{person}{Matthias Nie{\ss}ner}, {and} \bibinfo{person}{Christian Theobalt}.} \bibinfo{year}{2018}\natexlab{}.
\newblock \showarticletitle{State of the art on monocular 3D face reconstruction, tracking, and applications}. In \bibinfo{booktitle}{\emph{Computer graphics forum}}, Vol.~\bibinfo{volume}{37}. Wiley Online Library, \bibinfo{pages}{523--550}.
\newblock


\end{thebibliography}

\end{document}